\documentclass{article}
\usepackage{cite}
\usepackage{acra}
\usepackage{graphicx}
\usepackage{multirow}
\usepackage{adjustbox}
\usepackage{amsmath}
\usepackage{url}
\usepackage{hyperref}
\usepackage{array}
\usepackage{enumitem}

\title{Pallet Detection And Localisation From Synthetic Data}
\author{Henri Mueller*, Yechan Kim*, Trevor Gee, Mahla Nejati** \\ Centre for Automation and Robotic Engineering Science \\
The University of Auckland, New Zealand \\ 
*\{hmue642, ykim370\}@aucklanduni.ac.nz, **m.nejati@auckland.ac.nz}

\begin{document}

\maketitle

\begin{abstract}
The global warehousing industry is experiencing rapid growth, with the market size projected to grow at an annual rate of 8.1\% from 2024 to 2030  \cite{grandview2021warehouse}. This expansion has led to a surge in demand for efficient pallet detection and localisation systems. While automation can significantly streamline warehouse operations, the development of such systems often requires extensive manual data annotation, with an average of 35 seconds per image, for a typical computer vision project.

This paper presents a novel approach to enhance pallet detection and localisation using purely synthetic data and geometric features derived from their side faces. By implementing a domain randomisation engine in Unity, the need for time-consuming manual annotation is eliminated while achieving high-performance results. The proposed method demonstrates a pallet detection performance of 0.995 mAP50 for single pallets on a real-world dataset. Additionally, an average position accuracy of less than 4.2 cm and an average rotation accuracy of 8.2° were achieved for pallets within a 5-meter range, with the pallet positioned head-on.

\end{abstract}

\section{Introduction}
Accurate pallet detection and localisation are essential for optimising operations in various industries, including logistics, manufacturing, and warehousing. This research focuses on developing an effective system for detecting and localising individual pallets using only passive sensors. The primary motivation for this approach is to reduce the cost and complexity associated with using active sensors, such as laser scanners.

Building upon the previous work of \cite{Naidoo2023} and \cite{Gann2023}, this study aims to further enhance pallet detection, localisation accuracy and efficiency. \cite{Naidoo2023} used an Region-based Convolutional Neural Network (R-CNN) trained on data generated in a warehouse environment designed in a game engine called Unity. They achieved a mAP50 of 0.86 for single pallets. However, their stacked and racked pallet detection was 0.05 and 0.21, respectively. \cite{Gann2023} aimed to improve this by gathering more data, improving the warehouse environment, and switching to a different CNN called YOLOv8 \cite{Jocher_Ultralytics_YOLO_2023}. They also experimented with a two-stage segmentation pipeline by employing YOLO + SAM. \cite{Gann2023} achieved mAP50 values of 0.71, 0.74, and 0.71 for single, stacked, and racked pallet configurations, showing a performance improvement.

While previous studies primarily focused on pallet detection (\cite{Gann2023} and \cite{Naidoo2023}, this research seeks to enhance pallet localisation. Despite numerous investigations into pallet localisation, many rely on active 3D sensors \cite{t3,Li2021,Xiao2017,Zhao2022,Arpenti2020,Lv2023,Rocha2023}. This can be undesirable in a busy warehouse environment due to the increased risk of sensor interference. A passive RGB camera will be used to address this issue, which will also be more affordable. The system aims to determine a pallet's 3D position relative to the camera.

\begin{figure}[ht]
    \centering
    \includegraphics[width=\linewidth]{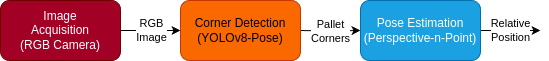}  
    \caption{Overview of the localisation System.}
    \label{fig:system}
\end{figure}

\begin{figure}[ht]
    \centering
    \includegraphics[width=\linewidth]{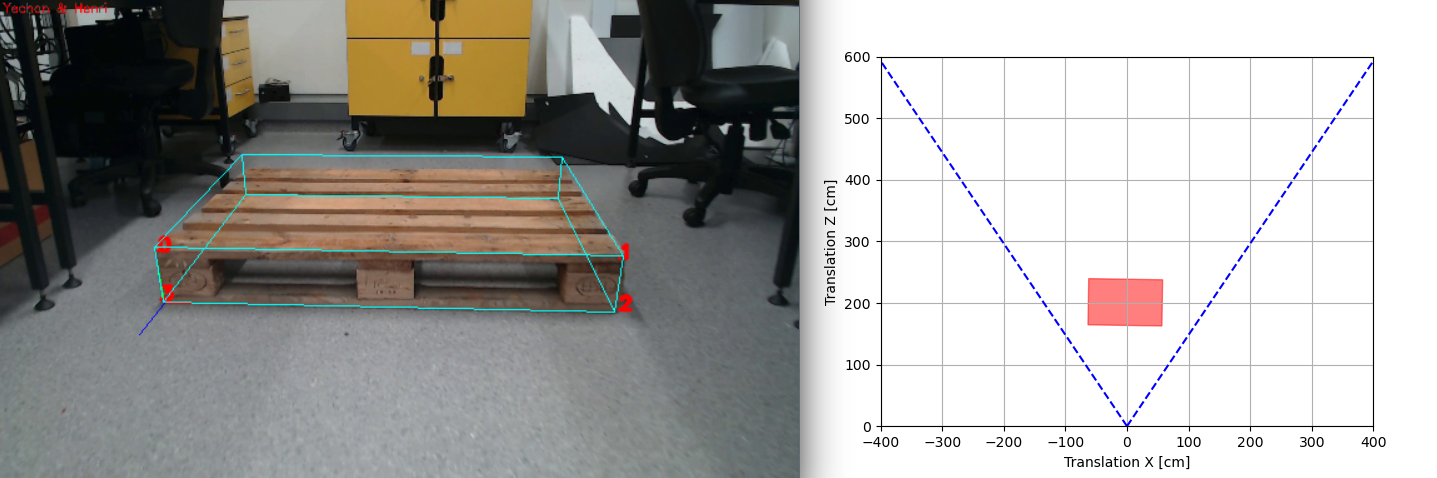}  
    \caption{Visualisation of a localised pallet using the proposed system}
    \label{fig:error_visulisation}
\end{figure}

A pipelined approach will be implemented, including image acquisition, corner detection using a keypoint detection Convolutional Neural Network (CNN), and pose estimation using the perspective-n-point algorithm. Figure \ref{fig:system} shows this pipeline, and Figure \ref{fig:error_visulisation} shows an example result of this approach. 

\section{Literature Review}
This section reviews advancements in pallet detection, focusing on sensors, detection methods, and localisation techniques. The review begins with an examination of various sensors and systems, followed by a discussion of detection and localisation methods, categorised into 2D and 3D sensors. The final part explores the use of synthetic data to improve these systems.

\subsection{Sensors and Systems}
Pallet interaction is inherently a closed-loop challenge, meaning that for an autonomous vehicle to engage with a pallet, it must:
\begin{enumerate}[label=\alph*)]
  \item Correctly identify the pallet's location
  \item Adjust its path accordingly
\end{enumerate}
This necessitates sensors with high resolution capable of detecting pallets accurately. Consequently, traditional ranging systems like RADAR are often not suitable.

\subsubsection{Monocular Cameras}
Monocular cameras have experienced a significant increase in popularity for pallet detection \cite{Kesuma2023,Zaccaria2020,Syu2017,Manurung2022,Bibob2020,Kita2019}, mainly attributable to their affordability and the progress in computer vision. Metal-oxide-semiconductor (CMOS) cameras capture photons to produce high-quality digital images. When integrated with CNNs, they show excellent performance in detecting objects within image frames \cite{Kesuma2023,Zaccaria2020,Manurung2022,Bibob2020}. These cameras are also referred to as Red-Green-Blue (RGB) cameras. 

A significant advantage of employing CMOS cameras in autonomous vehicles is that they are passive sensors. This is beneficial because they do not emit any signals that could potentially interfere with other sensors on the vehicles.

Despite their widespread use, these cameras do come with limitations. They lack the capability to retrieve 3D information, confining them to 2D data. Nevertheless, 3D information can be retrieved using geometric features and camera calibration. There are ongoing efforts underway to enhance detection algorithms to enable spatial estimation \cite{Bibob2020}. CMOS cameras struggle in low-light conditions \cite{Gann2023}, resulting in grainy images that may lead to misidentification and reduced system effectiveness.

\subsubsection{Depth Cameras}
Depth cameras have emerged as the preferred method for pallet detection and localisation in recent years, primarily due to their ability to gather 3D spatial data affordably \cite{Shao2022,Iinuma2021,t3,Li2021,Xiao2017,Zhao2022,Arpenti2020,Lv2023,Rocha2023,Jia2021}. This depth information enables a more accurate estimation of pallet positions.

These cameras can operate on various principles such as stereo vision, time-of-flight, and structured light. Additionally, most depth cameras incorporate an integrated CMOS camera, enabling the capture of depth and RGB images. Such sensor combinations are commonly known as Red-Green-Blue-Depth (RGB-D) cameras.

While depth cameras offer a significant advantage in providing 3D information, their accuracy can degrade with distance. Additionally, the calibration process necessary for accurate measurements can be time-consuming and costly. Additionally, they can struggle under certain conditions like bright sunlight or highly reflective objects

\subsubsection{Laser Scanners}
Laser Scanners, such as LiDAR, utilise laser beams to measure distances to objects and create detailed 3D representations of the surrounding environment. These systems emit laser pulses and measure the time the light returns after hitting objects, allowing for distance calculations. LiDAR scanners can generate highly accurate point cloud data, which can be used to detect pallet shapes \cite{Mohamed2018,phdthesis}. Despite their effectiveness, LiDAR systems can be expensive and are sensitive to environmental conditions such as fog and precipitation.

\subsection{Pallet Detection}
The initial stage in enabling an autonomous system to interact with a pallet is identification. This task can be accomplished using two main methods, neural networks \cite{Kesuma2023,phdthesis,Manurung2022,Latini2023,Wang2023,Tsiogas2021}, and feature detection \cite{Shao2022,Cai2022,Arpenti2020,Kesuma2023,Xiao2017,Zhao2022,Kita2019,Arpenti2020}. The information that they can process can be broken down into 2D data and 3D data. Where 2D data lacks depth information, like an RGB image.

\subsubsection{From 2D Information}

2D information from RGB images is widely used due to the ease of acquiring visual data from cameras \cite{Kesuma2023,phdthesis,Manurung2022,Latini2023,Wang2023,Tsiogas2021,Cai2022}. Methods based on CNNs, such as the YOLO (You Only Look Once) family of models, are particularly effective for detecting pallets in real-time \cite{Jocher_Ultralytics_YOLO_2023}. YOLO models excel at identifying objects by performing classification, which is important in dynamic environments like warehouses, where quick decisions are needed \cite{Manurung2022}. 


Traditional object detection methods often utilised algorithms like AdaBoost, which rely on manually extracted features for classification tasks \cite{Goswami2022}. However, these methods often struggle with accuracy in cluttered environments and require extensive manual feature extraction. In contrast, deep learning models like YOLO and Faster R-CNN have become more common because they automatically extract features from data and perform better in challenging scenarios \cite{Goswami2022}.

\subsubsection{From 3D Information}
There are various methods for detecting pallets from 3D information. 

One approach is to use the point cloud from a laser scanner, which is slightly mounted above the floor. The top-down point cloud would then be fed to a CNN, which would identify the pallet's structure legs \cite{phdthesis}. This method proved to be very promising, with a 98.57\% detection rate. However, the author also mentioned that this number would decrease as more objects entered the scene.

Another approach is to fuse the data from an RGB camera with that from a Depth camera \cite{Wang2023}. This system uses a deep neural network to process the RGB image and depth map and identify and locate pallets. The system had a mAP of 79.2, compared to YOLOv5 of 83.5. It should be noted that this system also performs localisation. The drawback of this approach is the use of an active sensor.

\subsection{Pallet Localisation}
Pallet localisation presents a significantly greater challenge than detection, as it requires more extensive and precise data to determine the orientation and distance of the pallet accurately. Unlike detection, 3D sensors are most used for this as the spatial information makes it much easier to determine position \cite{Wang2023,t3,Li2021,Xiao2017,Zhao2022,Arpenti2020,Lv2023,Rocha2023,Jia2021}. However 2D approaches exist as well \cite{Bibob2020,Kita2019}. 

\subsubsection{From 2D Information}
Estimating the position based on an RGB image can be very challenging. However, this can be achieved using a neural network that outputs orientation information. This neural network was called Deep Object Pose Estimation. This system proved to have an average position error of less than 20 cm \cite{Bibob2020}. It should be noted that they trained the model on synthetic data, which tends to be less accurate. The tests were conducted on an empty floor with a pallet in the middle. The author also noted that an increase in scene complexity reduced the effectiveness. 

Another method for retrieving the 6D position from a 2D image was using a high FOV camera as described in \cite{Kita2019}. This has been done by re-projecting wide-angle views onto a 3D plane. The system's accuracy proved to be good, 19mm in x, 0.7mm in y, and 100.1mm in z at 4868mm. One drawback of this system is that the camera needs to be angled away from the pallet, and the vehicle needs to move in an offset path to localise the pallet.

\subsubsection{From 3D Information}
The most common method of determining localisation is using an RGB-D camera \cite{t3,Li2021,Xiao2017,Zhao2022,Arpenti2020,Lv2023,Rocha2023}. The generated point cloud was first cleaned using random sample consensus (RANSAC), after which the processed point cloud was passed onto various algorithms. 

Another method used two neural networks, one for the RGB image and one for the point cloud. Using YOLOv8 \cite{t3}, the region of interest (the area where a pallet sits) is determined. This area of the point cloud was passed into PointNet, which gave it a 6D pose. This system had a stated accuracy of 0.77 - 0.83 AP.

Another method is by processing the point cloud data using template matching. The point cloud was first segmented using a hybrid region growing algorithm \cite{Xiao2017}. Then, a known 3D template of the pallet is used to compare the segmented regions. The template is then moved in the Z axis to give the lowest error, thereby calculating the pose. As the system was used to guide the forks, the error was given at the point where the forks were inserted into the pallet; this was 0.03m.

Another approach is to segment the image first and then perform template matching on the segmented image \cite{Zhao2022}. First, it classified every pixel of a colour image by its colour feature, generating a category matrix. Then, a labelled template was constructed, containing information about the load, pallet, and ground categories. Third, the category matrix and template are compressed and matched to the pallet's region. Finally, the pallet's position was extracted using information from its feet. The maximum distance and angle estimation error was -101.1 mm and 6.07° in a range from 1 m to 4 m and an angle of within ±25°.

\subsection{Synthetic Data}
Training large deep learning models requires extensive labelled datasets, which can be expensive and time-consuming to collect manually \cite{AdamZewe2022,10042887}. Synthetic data generation provides an alternative by automating the creation and annotation of data, significantly reducing the cost and effort involved \cite{AdamZewe2022}. This approach allows models to be trained with minimal real-world data or, in some cases, no real data at all, which is particularly useful in environments where data collection is challenging.

Several platforms, such as Unity, Isaac Sim, and BlenderProc, have been used for generating synthetic data in various applications \cite{Gann2023,Naidoo2023,9987772}. Unity, for example, has demonstrated effectiveness in warehouse environments, particularly in pallet detection \cite{Gann2023,Naidoo2023}. 
In \cite{Gann2023}, they generated complex pallet configurations, such as stacked and racked pallets, resulting in a 69\% increase in mAP50 for stacked pallets \cite{Gann2023}.

Generating this data can be done in different ways, one way, is called domain randomisation, which randomly introduces controlled variability in training data by altering environmental factors like lighting, textures, and object placement; it is widely used to enhance the generalisation of models trained on synthetic datasets \cite{StefanGrushko2023}. \cite{StefanGrushko2023} demonstrated the effectiveness of domain randomisation in generating synthetic RGB-D datasets for hand localisation in cluttered industrial environments. 

Despite its advantages, synthetic data generation has limitations. While synthetic datasets can capture many real-world variations, they may not always reflect the full complexity of real environments \cite{Hello123}. Domain randomisation requires larger synthetic datasets to match the performance of real data, and models may still need fine-tuning with real-world data to bridge the "reality gap" fully \cite{StefanGrushko2023}. Nevertheless, \cite{10042887} reported an 11.8\% improvement in object detection performance when using synthetic data alone and a 37\% improvement when all randomisation techniques were applied compared to baseline models. These experiments were conducted with objects resembling what would be found in assembly plants, although pallets were not part of the dataset.

\subsection{Research Gaps}
While there has been exhaustive research into pallet detection and localisation, some research gaps remain.

One gap found is the ability to accurately and rapidly localise a pallet based on monocular RGB images. This area is of interest as passive sensors are preferred over active sensors. While papers such as \cite{Bibob2020} and \cite{Kita2019} only used RGB images, \cite{Kita2019} requires extra motion, which may not be practical to perform. Additionally, no research has been done utilising keypoint tracking for pallet pose estimation. 


This study aims at investigating this problem.

\section{Methods}
This section outlines the methodology employed for the pallet localisation task. The study utilised the YOLOv8 model for its object detection and pose estimation capabilities. The model was trained using a synthetic dataset generated in Unity, which included various pallet orientations and lighting conditions.

The localisation process consisted of three key stages: image acquisition, corner detection, and pose estimation. Corner detection was performed using keypoint tracking, a technique adapted from human pose estimation. Finally, the pallet's pose was calculated using the PnP algorithm.

\subsection{YOLOv8}
YOLOv8 is a state-of-the-art computer vision model developed by Ultralytics, widely recognised for its high performance in object detection, classification, and segmentation tasks \cite{Jocher_Ultralytics_YOLO_2023}. In addition to these core tasks, YOLOv8 has also demonstrated high performance in pose estimation, a feature that involves detecting key points on objects or individuals within images or video frames. This functionality is essential for tasks requiring precise localisation, such as pallet detection, where the keypoints of pallet corners are required to determine orientation and position.

Although newer versions like YOLOv9 \cite{wang2024yolov9learningwantlearn} and YOLOv10 \cite{wang2024yolov10realtimeendtoendobject} have been developed with improvements in areas such as instance segmentation and object detection, they do not offer keypoint tracking as of the time of testing. Therefore, despite the advancements in YOLOv9 and YOLOv10, YOLOv8 remains the preferred model for this study due to its pose estimation feature \cite{Jocher_Ultralytics_YOLO_2023}.

\subsection{Synthetic Data Generation}
As in \cite{Naidoo2023} and \cite{Gann2023}, Unity was used to generate the synthetic dataset. However, unlike the previous two studies, a domain randomisation environment was developed, as this approach was only briefly explored by \cite{Gann2023}. Although software specifically made for domain randomisation exists, such as Nvidia's Issac Sim, it cannot produce datasets for keypoint tracking.

\begin{figure}[ht]
    \centering
    \includegraphics[width=\linewidth]{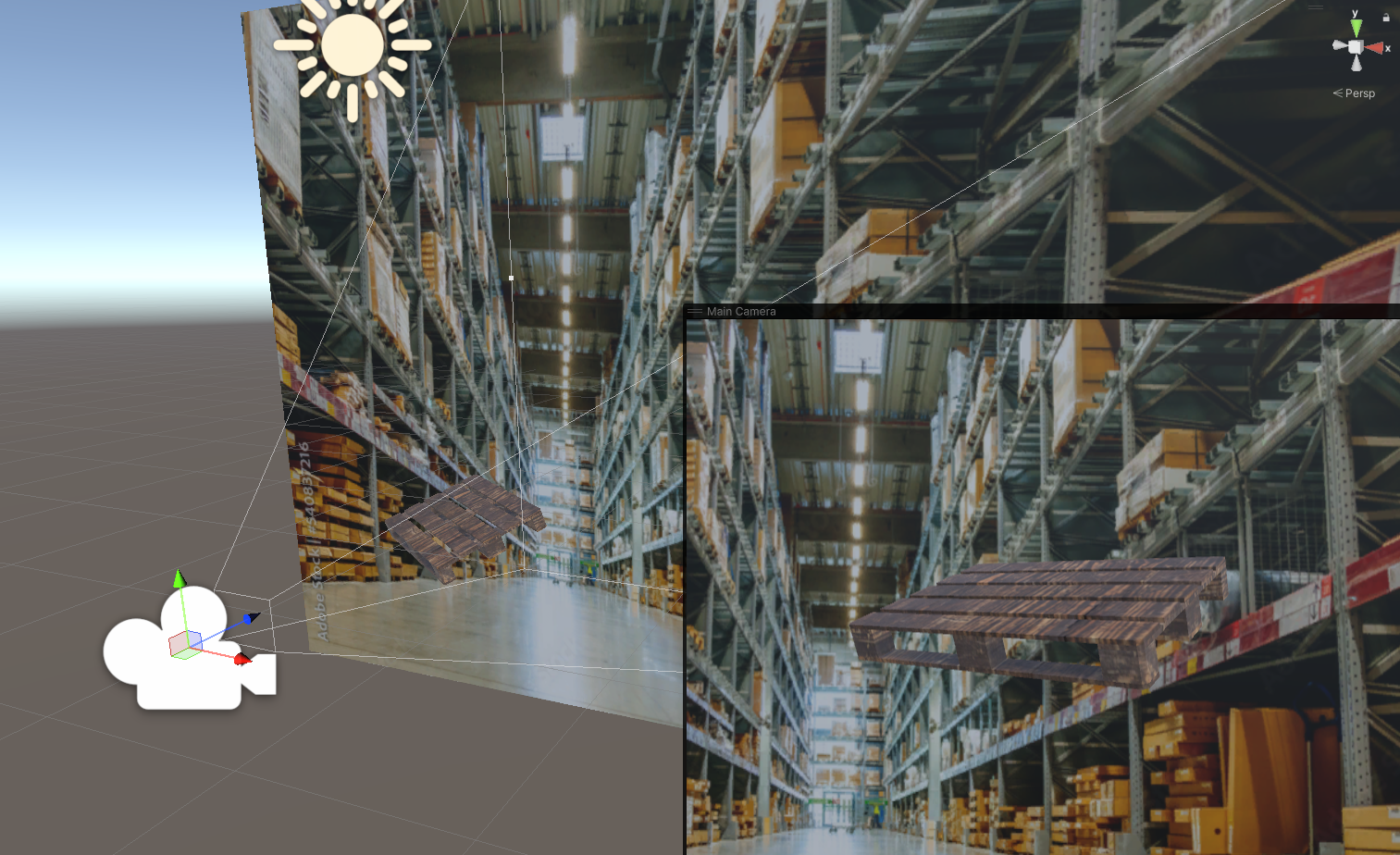}  
    \caption{Domain Randomisation Scene.}
    \label{fig:pallet_unity}
\end{figure}

The generated dataset aimed to capture the front (1200 mm) and side (800 mm) views of the pallets, with an equal split of views. This can be seen in Figure \ref{fig:error_vis}. The two sides require separate distinctions due to the differences in dimension. For each frame, domain randomisation was applied to adjust the intensity of the scene's primary light source, pallet texture and position (x, y, z axes), and pallet orientation. Previous studies did not take into consideration varying pallet textures. However, in the real world, pallets can have different levels of wear. An example of the Unity environment can be seen in Figure \ref{fig:pallet_unity}.

Five YOLOv8 model variants were trained. Nano, small, medium, large, and extra-large. These variants differ in model size and complexity, where trade-offs between accuracy and computational efficiency. Table \ref{yolomodels} shows more detail between models.

The metric mAP (mean Average Precision) refers to the accuracy of object detection systems across multiple recall levels, providing a single performance score. Params refers to the number of parameters in a model, indicating its complexity and capacity to learn from data. FLOPs (Floating Point Operations per Second) measure the computational efficiency of a model.

\begin{table}[ht]
\centering
\begin{tabular}{|c|c|c|c|}
\hline
\textbf{Model} & \textbf{mAP50-95} & \textbf{Params} & \textbf{FLOPs} \\
\hline
YOLOv8n & 0.373 & 3.2 M  & 8.7 G  \\
\hline
YOLOv8s & 0.449 & 11.2 M & 28.6 G \\
\hline
YOLOv8m & 0.502 & 25.9 M & 78.9 G \\
\hline
YOLOv8l & 0.529 & 43.7 M & 165.2 G \\
\hline
YOLOv8x & 0.539 & 68.2 M & 257.8 G \\
\hline
\end{tabular}
\caption{YOLO Models Comparison. \protect\cite{Jocher_Ultralytics_YOLO_2023}}
\label{yolomodels}
\end{table}

This study utilised varying sizes of the synthetic training dataset. These were 7,500 images (7.5k), 15,000 images (15k), and 30,000 images (30k). This was done to compare how different dataset sizes would impact the performance of the models. Generating the training data proved to be relatively quick. The Unity domain randomiser took 8m 41s to generate the 30,000 images with labels on an Intel i7-10600KF. If using the 35 seconds per image annotation given from \cite{papadopoulos2017extremeclickingefficientobject}, this process could take more than 290 hours. This shows the value that synthetic data generation can bring.

\subsection{Localisation}

To calculate the pose of a pallet using geometric features extracted from an image, two assumptions must be made: the pallet is correctly identified, and the pallet face dimensions are known. For the first assumption, testing was performed on the Euro-pallet type 1, and the model was trained exclusively on this pallet type. Figure \ref{fig:pallet_unity} shows a 3D model inside of Unity of this type of pallet. For the second assumption, since only one type of pallet was used, its face dimensions are known to be 1200x144 mm (length/height) and 800x144 mm (width/height).

The system operated as a pipeline with three stages: image acquisition, corner detection, and pose estimation, as shown in Figure \ref{fig:system}. In the image acquisition stage, an image was captured, pre-processed, and then fed into the neural network. The corner detection stage involved identifying the pallet's corners through keypoint tracking. Finally, the pose estimation stage calculates the pallet's pose based on the identified keypoints.

\subsection{Corner Detection}
To identify the corners of a pallet's face, keypoint tracking was used, a method first implemented by \cite{majiyolopose} and integrated with YOLOv8. Originally designed for joint position estimation to determine human poses, this system was adapted to identify pallet corners.

The keypoint system operated by associating each detected pallet face with an anchor box that contains both the bounding box coordinates and the full 2D pose information, which includes the locations of the keypoints. The model regresses the coordinates of 4 keypoints for each pallet relative to the anchor's centre. This allows for accurate localisation of keypoints. The use of Object Keypoint Similarity as a loss function optimises the accuracy of keypoint predictions by weighing the importance of each keypoint based on its visibility and relevance.

\subsection{Pose Calculation}
The first step in pose estimation is to determine the camera's intrinsic matrix. The camera used was a Logitech C920 HD Pro Webcam, as it allows for manual focus along with a high image resolution. A ChArUco board pattern was used to calibrate the camera, as it is much more accurate than a traditional chessboard pattern, as stated by \cite{bradski2008learning}. Each square on the board provides known 3D points corresponding to detected 2D points in the images. By matching these correspondences, the camera's intrinsic parameters, such as focal length and optical centre, can be estimated. This was done by minimising the re-projection error between the observed and predicted image points.

The points found in the corner detection step are used to estimate the pose of the pallet. The translation and rotation of a pallet in 3D space relative to a camera are estimated using the Perspective-n-Point (PnP) algorithm \cite{zhouPnP}. The process begins by matching a set of 2D points in an image (corner points) to their corresponding 3D points in the real world (where z=0). The algorithm then uses this information, along with the camera's intrinsic parameters, to calculate the pallet's position and orientation.

To calculate the error of the system, an ArUco marker was placed on top of the pallet to be used as a ground truth.

\section{Results}
This section will cover the results of the detection and localisation system. The performance evaluation was conducted on a dataset consisting of real-world pallet detection scenes. The test dataset used in this evaluation consisted of 60 manually annotated images extracted from six distinct video sequences covering multiple pallet orientations. 

To generate the real-world dataset, a series of videos were taken, of which 60 frames were extracted and then manually labelled. To make the videos, a pallet was fixed on the ground, with the camera base moving towards it. The camera base started 5 meters from the pallet and moved forward while recording the video. This process was repeated with the pallet rotated 35° to the left and right for each side.

\subsection{Pallet Detection}
Following the approach of \cite{Gann2023}, the pallet detection comparison included models of different sizes to examine if they have different performances. 

The training and validation data was split in an 80-20 ratio for all models. The hyperparameters used were a batch size of 3 as the limiting factor was the Nvidia RTX 3080 10GB GPU's memory. The other hyperparameters can be seen in the Table \ref{tab:hyperparameters}.

\begin{table}[ht]
    \centering
    \begin{tabular}{|c|c|c|c|c|}
    \hline
    Image Size & Initial LR & Momentum & Decay \\
    \hline
    640x384 & 0.01 & 0.937 & 0.0005 \\
    \hline
    \end{tabular}
    \caption{Hyperparameters used for Models}
    \label{tab:hyperparameters}
\end{table}

The results presented in Table \ref{tab:model_comparison_table} highlight the detection accuracy for each model, measured using mAP50 and mAP50-95, and the Keypoint Mean Error (KME). KME is measured in pixel error $\times 10^{-6}$.

\begin{table}[ht]
    \centering
    \begin{tabular}{|l|c|c|c|}
        \hline
        \textbf{Model}     & \textbf{mAP50} & \textbf{mAP50-95} & \textbf{KME} \\ \hline\hline
        YOLOv8n 30k        & 0.971          & 0.54              & 11012.27     \\ \hline
        YOLOv8n 15k        & 0.988          & 0.547             & 8740.43      \\ \hline
        YOLOv8n 7.5k       & 0.988          & 0.557             & 4683.51   
        
        \\ \hline\hline
        YOLOv8s 30k        & 0.985          & 0.555             & 3318.03      \\ \hline
        YOLOv8s 15k        & 0.994          & 0.561             & 3666.31      \\ \hline
        YOLOv8s 7.5k       & 0.987          & 0.540             & 4718.57      
        
        \\ \hline\hline
        YOLOv8m 30k        & 0.973          & 0.562             & 8850.03      \\ \hline
        YOLOv8m 15k        & 0.962          & 0.579             & 4541.41      \\ \hline
        YOLOv8m 7.5k       & 0.964          & 0.546             & 6488.15      

        \\ \hline\hline
        YOLOv8l 30k        & 0.990          & 0.597             & 9297.20      \\ \hline
        YOLOv8l 15k        & 0.995          & 0.599             & 3463.09      \\ \hline
        YOLOv8l 7.5k       & 0.971          & 0.574             & 4567.77      

        \\ \hline\hline
        YOLOv8x 30k        & 0.995          & 0.597             & 8703.45      \\ \hline
        YOLOv8x 15k        & 0.995          & 0.588             & 2960.90      \\ \hline
        YOLOv8x 7.5k       & 0.960          & 0.569             & 3302.36      \\ \hline
    \end{tabular}
    \caption{\textbf{Performance of YOLOv8 models with different training sizes.}}
    \label{tab:model_comparison_table}
\end{table}

It can be seen that different dataset sizes significantly affect the models' performance. Larger models like YOLOv8l and YOLOv8x achieved near-identical performance, with the highest mAP50 value of 0.995 on the 30k dataset, while smaller models like YOLOv8n achieved a lower mAP50 of 0.971. A similar trend can be observed for the mAP50-95 metric, where YOLOv8l and YOLOv8x have the highest performance, achieving 0.597 and 0.599 on the 30k datasets. The smaller models, particularly YOLOv8n, perform significantly worse under these conditions, dropping to 0.54 mAP50-95.

To evaluate the accuracy of pallet corner detection, the Keypoint Mean Error metric was used to provide a linear pairwise comparison between each ground-truth corner and the predicted corner coordinates. The ground-truth corners were manually annotated on each test image. Since pallet detection requires accurate corner localisation, this metric directly measures the models' ability to predict these keypoints. By averaging the pairwise distance between predicted and actual keypoints provides insights into spatial precision. This method captures details that metrics like mAP50 or mAP50-95 might not reflect as effectively.

\subsection{Pallet Localisation}

\begin{figure}[ht]
    \centering
    \includegraphics[width=\linewidth]{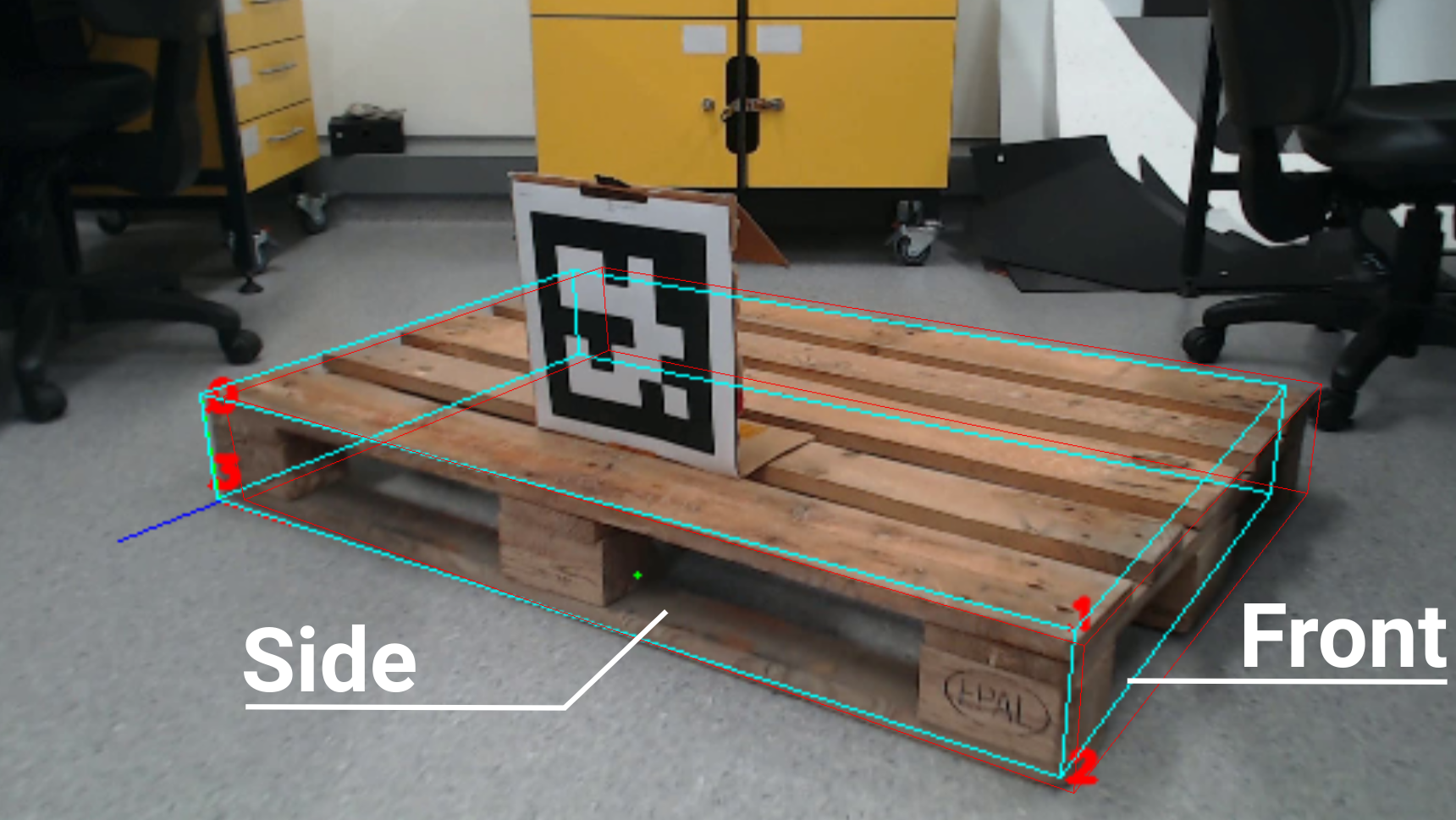}  
    \caption{Localisation Error Visualisation.}
    \label{fig:error_vis}
\end{figure}

A baseline system for ground truth comparison is required to measure the accuracy of the localisation system. Initially, a high-resolution 3D camera (Zivid One+ Medium) was considered the reference. However, it proved unsuitable due to the RGB image's narrow field of view, which could not fully capture the pallet within the sensor's maximum range. Instead, an ArUco marker was placed on top of the pallet to assess position accuracy, as shown in Figure \ref{fig:error_vis}. While this approach has limitations since its accuracy depends on the ArUco marker localisation algorithm — it is assumed that the error introduced by this algorithm is significantly smaller than the error of the pallet localisation system. According to \cite{arucomarkeracc} ArUco markers have a 98\% localisation accuracy.

\begin{figure}[ht]
    \centering
    \includegraphics[width=\linewidth]{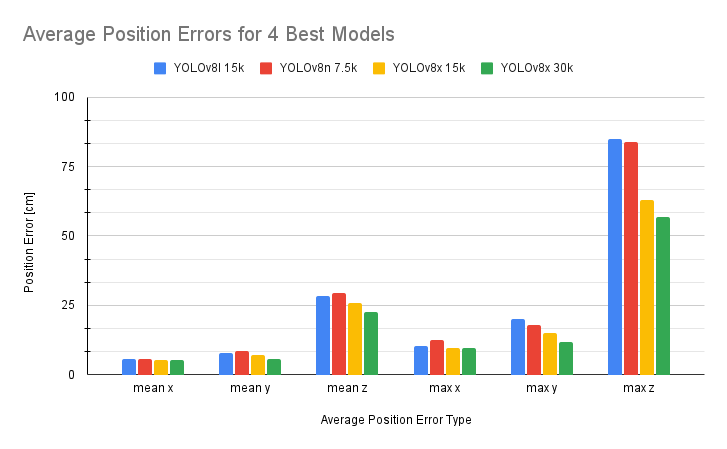}  
    \caption{Comparative Position Error}
    \label{fig:error_comparison_pos}
\end{figure}

\begin{figure}[ht]
    \centering
    \includegraphics[width=\linewidth]{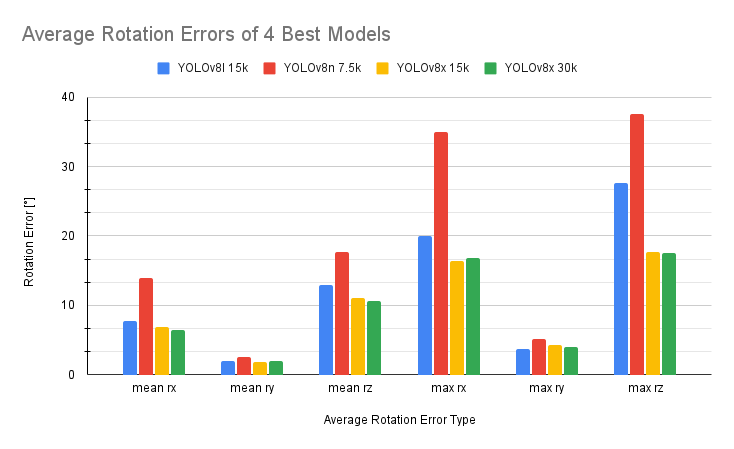}  
    \caption{Comparative Rotation Error}
    \label{fig:error_comparison_rot}
\end{figure}

Based on the results in Table \ref{tab:model_comparison_table}, the following models were investigated for pose estimation: YOLOv8l 15k, YOLOv8n 7.5k, YOLOv8x 15k, and YOLOv8x 30k. The performance comparison can be seen in Figure \ref{fig:error_comparison_pos} and Figure \ref{fig:error_comparison_rot}.

Using the results from Figure \ref{fig:error_comparison_pos} and Figure \ref{fig:error_comparison_rot}, it can be determined that the YOLOv8x model with 30,000 training images seems to perform the best.

The following provides a more detailed look at the error metrics. The model used for the subsequent tests is YOLOv8x 30k. This model was chosen as it had the highest testing score for both detection and keypoints. It also showed the lowest position and rotation errors.

A more detailed overview of each test can be seen in Table \ref{tab:yolov8x30k} where 35L/R refers to the pallet being rotated 35° left or right. "Front" and "Side" refers to the pallet's side faces, as seen in Figure \ref{fig:error_vis}. Average translation errors (x, y, z) are given in centimetres, while average rotation errors (rx, ry, rz) are in degrees. 

\begin{table}[ht]
    \centering
    \begin{tabular}{|l|c|c|c|c|c|c|}
        \hline
        \textbf{Side} & \textbf{x} & \textbf{y} & \textbf{z} & \textbf{rx} & \textbf{ry} & \textbf{rz} \\
        \hline
        Side         & 0.8   & -2.6  & -1.9  & -16.9  & -1.1   & -4.2   \\
        Side 35L     & -8.3  & -11.8 & 56.7  & 8.4    & -3.9   & 17.6   \\
        Side 35R     & -9.7  & -10.0 & 34.4  & -0.7   & -1.0   & -17.3  \\
        Front        & -4.2  & -1.5  & -2.3  & -8.2   & -0.6   & -6.7   \\
        Front 35L    & -1.5  & -2.4  & 21.9  & -3.1   & 3.5    & 9.3    \\
        Front 35R    & -7.3  & -5.2  & 19.1  & 1.5    & -1.3   & -8.4   \\
        \hline
    \end{tabular}
    \caption{Average Errors for Translation (x, y, z) and Rotation (rx, ry, rz)}
    \label{tab:yolov8x30k}
\end{table}

\begin{figure}[ht]
    \centering
    \includegraphics[width=\linewidth]{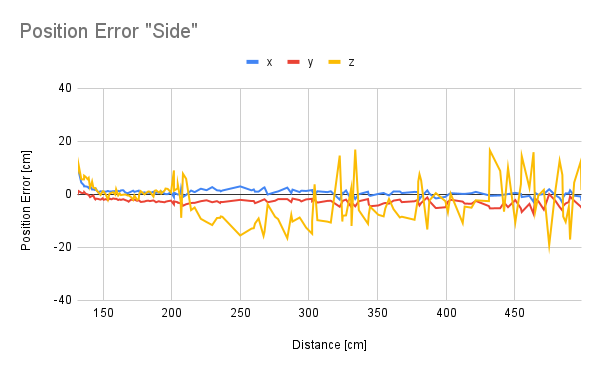}  
    \caption{Position Error for Side}
    \label{fig:error_posgraph}
\end{figure}

\begin{figure}[ht]
    \centering
    \includegraphics[width=\linewidth]{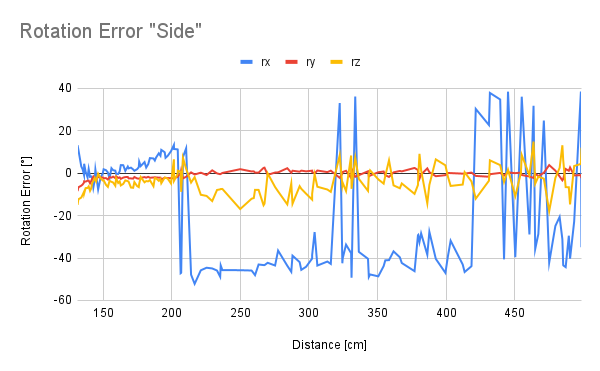}  
    \caption{Rotation Error for Side}
    \label{fig:error_rotgraph}
\end{figure}

Figures \ref{fig:error_posgraph} and \ref{fig:error_rotgraph} give a more detailed breakdown of the errors for the pallet "Side" configuration up to 5m. The other configurations had similar performance. However, due to space efficiency they are omitted.

One issue with this system can be seen in Figure \ref{fig:error_rotgraph}, where the rotation around the X-axis can be irregular.

\section{Discussion}
As observed in our results (Table \ref{tab:model_comparison_table}), there is a clear trend where larger models, such as YOLOv8l and YOLOv8x, consistently outperform smaller models like YOLOv8n and YOLOv8s in terms of both mAP50 and mAP50-95 scores. This improvement is attributed to the increased complexity in larger models  \cite{Jocher_Ultralytics_YOLO_2023}. However, this also means that they require more computational resources and are more prone to overfitting if not properly regularised \cite{supp1}.

Previous studies \cite{Gann2023} achieved a mAP50 of 0.71, where this discrepancy is mostly due to two reasons. This study focused on only using one type of pallet, so there was less generalisation. Secondly, this study introduced variable pallet textures, improving the detection of non-uniformly coloured pallets.

Generally, larger models achieve a lower KME score, indicating better keypoint localisation accuracy. However, YOLOv8s trained on the 30k dataset achieved a lower KME than the larger YOLOv8l on the same dataset. This suggests that smaller models can sometimes surpass larger ones in keypoint localisation when provided with sufficient training data. 

For the localisation, it can be seen from Table \ref{tab:yolov8x30k} that the "Side" configuration has the smallest translation error in the x-direction (0.8 cm) but significant rotation errors, particularly in the x-axis (-16.9°). In contrast, the "Side 35L" and "Side 35R" configurations exhibit the largest translation errors, especially in the z-direction (56.7 cm and 34.4 cm, respectively). The "Front" configuration shows relatively small errors in both translation and rotation, while "Front 35L" and "Front 35R" have moderate translation errors, with rotation errors varying across different axes. The "Front" configuration showed the best overall performance, with a maximum average translation error of 4.2 cm and rotation error of 8.2°. Overall, side configurations tend to have larger translation errors than the front ones. This is comparable to \cite{Bibob2020}, who achieved a sub-20 cm position accuracy.

\begin{figure}[ht]
    \centering
    \includegraphics[width=\linewidth]{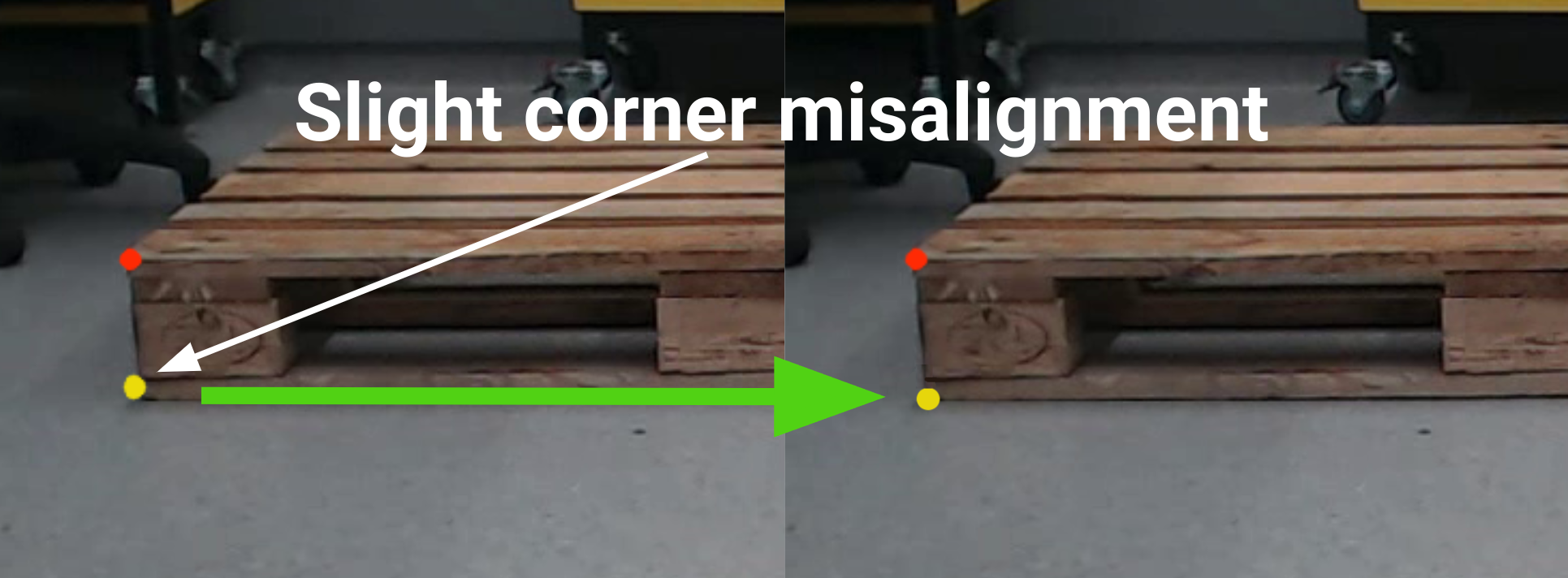}  
    \caption{Example of the bottom left corner being slightly misaligned.}
    \label{fig:bad_corners}
\end{figure}

The high x-axis rotation error occurs because the system misidentifies the pallet's corners. As shown in Figure \ref{fig:bad_corners}, the bottom corners are detected higher than they should be, reducing the height-to-width ratio. This makes the pallet appear as if it is being viewed from an upper or lower angle, leading to the error.

The primary limitation in assessing accuracy was the lack of a reliable ground truth for comparison. Another limitation was the availability of only a single pallet, which restricted the ability to test the system in stacked or racked pallet configurations. 

\section{Conclusions and Future Work}
The primary research goal of this project was to improve the detection and localisation of pallets using only synthetic data and to test the system by evaluating it on a real-world dataset.

The pallet detection results demonstrate that the YOLOv8x model with 15k images is the most suitable for this problem. When tested on real-world data, the model achieved a mAP50 of 0.995 and a mAP50-95 of 0.588, showing its strong performance in identifying pallets under various conditions.

The localisation results show that estimating a pallet's rough position and rotation is possible using a set of geometric features. The system achieved a total mean position error of 3.6 cm and a total mean rotation error of -1.83°. 

Key limitations, such as using only a single pallet and the absence of reliable ground truth for comparison, impacted the full evaluation of the system's capabilities. These factors should be addressed in future work to enhance both the system's effectiveness and the accuracy of its results, especially in scenarios involving multiple pallets or larger distances.

Other future work will include finding a better ground truth method to test the localisation system more accurately. Adding more pallet types to the dataset to distinguish from non-Euro-type pallets. Real-world tests in a warehouse will be performed to identify further problems. Additionally, more research should be conducted for the detection of wrapped or packaged pallets.

\section*{Acknowledgements}
We extend our sincere gratitude to Crown Equipment Limited, with special thanks to Sian Phillips and Amirali Pourgolmohammadgolshani, for sponsoring this project and providing the Euro Pallet along with additional images.

\bibliography{t2}
\bibliographystyle{apalike}

\end{document}